\documentclass[letterpaper, 10 pt, conference]{ieeeconf}  %

\IEEEoverridecommandlockouts                              %

\usepackage{graphics} %
\usepackage{epsfig} %
\usepackage{times} %
\usepackage{amsmath} %
\usepackage{amssymb}  %

\usepackage{algorithmicx, algpseudocode, algorithm}
\usepackage{booktabs}
\usepackage{cite}
\usepackage[font=small,labelfont=bf]{caption}
\usepackage{subcaption}
\usepackage{dcolumn}

\usepackage{bm}

\usepackage[printonlyused]{acronym}
\acrodef{AC}{Actor-Critic}
\acrodef{AD}{algorithmic differentiation}
\acrodef{AGV}{autonomous guided vehicle}
\acrodef{AGR}{autonomous ground robot}
\acrodef{AHRS}{attitude heading reference system}
\acrodef{AOW}{ANYmal On Wheels}
\acrodef{ASL}{Autonomous Systems Lab}
\acrodef{CAD}{computer-aided design}
\acrodef{CARE}{continuous time algebraic Ricatti equation}
\acrodef{COM}{center of mass}
\acrodefplural{COM}{centers of mass}
\acrodef{DAE}{differential algebraic equation}
\acrodef{DARE}{discrete time algebraic Ricatti equation}
\acrodef{DDP}{differential dynamic programming}
\acrodef{DOF}{degree of freedom}
\acrodefplural{DOF}{degrees of freedom}
\acrodef{EKF}{extended Kalman filter}
\acrodef{EOM}{equations of motion}
\acrodef{ETHZ}{Swiss Federal Institute of Technology Zurich}
\acrodef{FOV}{Field of View}
\acrodef{IL}{Imitation Learning}
\acrodef{IMU}{inertial measurement unit}
\acrodef{KF}{Kalman filter}
\acrodef{KKT}{Karush-Kuhn-Tucker}
\acrodef{LHS}{left-hand side}
\acrodef{LICQ}{linear independence constraint qualification}
\acrodef{LQR}{linear-quadratic regulator}
\acrodef{LOS}{line of support}
\acrodefplural{LOS}{lines of support}
\acrodef{MDP}{Markov Decision Process}
\acrodef{MLP}{Multi Layer Perceptron}
\acrodef{MPC}{model predictive control}
\acrodef{MSE}{Mean Squared Error}
\acrodef{NLP}{nonlinear program}
\acrodef{PD}{proportional-derivative}
\acrodef{PF}{particle filter}
\acrodef{PI}{proportional-integral}
\acrodef{PID}{proportional-integral-derivative}
\acrodef{PPO}{Proximal Policy Optimization}
\acrodef{PVT}{position-velocity-torque}
\acrodef{QP}{quadratic program}
\acrodef{RHS}{right-hand side}
\acrodef{RL}{Reinforcement Learning}
\acrodef{RND}{Random Network Distillation}
\acrodef{ROS}{Robot Operating System}
\acrodef{RSL}{Robotic Systems Lab}
\acrodef{SQP}{sequential quadratic programming}
\acrodef{TO}{trajectory optimization}
\acrodef{TV-LQR}{time-varying linear-quadratic regulator}
\acrodef{UKF}{unscented Kalman filter}
\acrodef{WIP}{wheeled inverted pendulum}
\acrodef{WBC}{whole-body control}
\acrodef{ZMP}{zero-moment point}
\acrodef{ZOH}{zero-order hold}

\usepackage{siunitx}

\usepackage{amssymb}

\usepackage{multirow}

\usepackage{cancel}

\usepackage{aligned-overset}

\usepackage{balance}

\usepackage{csquotes}

\usepackage{hyperref}

\RequirePackage{soul}

\usepackage[dvipsnames,svgnames,x11names,table]{xcolor}

\usepackage{soulutf8}
\soulregister\cite7
\soulregister\autoref7
\soulregister\fieref7
\soulregister\ref7
\soulregister\eqref7
\soulregister\ac7
\soulregister\footnote8
\soulregister\SI{1}

\usepackage{sidecap} %

\newcommand{\etal}{\textit{et al}. }
\newcommand{\ie}{\textit{i}.\textit{e}., }

\newcommand{\secref}[1]{Sec.~\ref{#1}}
\newcommand{\eqtref}[1]{Eq.~\ref{#1}}
\newcommand{\figref}[1]{Fig.~\ref{#1}}
\newcommand{\tabref}[1]{Table~\ref{#1}}

\usepackage{amsmath,amsfonts,bm}

\def\des{\textrm{des}}
\def\Rsp{{\mathbb{R}}}

\def\mdp{\mathcal{M}}
\def\S{\mathcal{S}}
\def\s{{s}}
\def\A{{\mathcal{A}}}
\def\a{{a}}

\def\r{{r}}
\def\T{{T}}

\newcommand{\grad}{\nabla}

\newcommand{\E}{\mathbb{E}}

\newcommand{\normtwo}[1]{\left\lVert#1\right\rVert_2}

\newcommand{\gradth}{\grad_{\theta}}

\newcommand{\Eb}[2]{\E_{#1}\left[#2\right]}

\newcommand{\pithet}{{\pi_{\theta}}}
\newcommand{\pithetk}{{\pi_{\theta_k}}}
\newcommand{\pithetkg}{{\pi_{\theta_k}^g}}

\newcommand{\loss}{\mathcal{L}}

\newcommand{\Apithetk}{A^{\pithetk}}
\newcommand{\Apithetkg}{A^{\pithetkg}}
\newcommand{\Api}{A^{\pi}}
\newcommand{\Vpi}{V^{\pi}}
\newcommand{\Qpi}{Q^{\pi}}

\usepackage{misc/onimage}

\tikzset{
    image label/.style={
        every node/.style={
            fill=black,
            text=white,
            font=\fontfamily{phv}\selectfont\small\bfseries,
            anchor=north east,
            xshift=-0.05cm,
            yshift=-0.05cm,
            at={(1,1)}
        }
    }
}

\tikzset{
    image label_tl/.style={
        every node/.style={
            fill=black,
            text=white,
            font=\fontfamily{phv}\selectfont\small\bfseries,
            anchor=north west,
            xshift=0.1cm,
            yshift=-0.1cm,
            at={(0,1)}
        }
    }
}

\tikzset{
    image label_tr/.style={
        every node/.style={
            fill=black,
            text=white,
            font=\fontfamily{phv}\selectfont\small\bfseries,
            anchor=north east,
            xshift=-0.1cm,
            yshift=-0.1cm,
            at={(1,1)}
        }
    }
}

\tikzset{
    image label_bl/.style={
        every node/.style={
            fill=black,
            text=white,
            font=\fontfamily{phv}\selectfont\small\bfseries,
            anchor=south west,
            xshift=0.1cm,
            yshift=0.1cm,
            at={(0,0)}
        }
    }
}

\tikzset{
    image label_br/.style={
        every node/.style={
            fill=black,
            text=white,
            font=\fontfamily{phv}\selectfont\small\bfseries,
            anchor=south east,
            xshift=-0.1cm,
            yshift=0.1cm,
            at={(1,0)}
        }
    }
}

\hypersetup{
    colorlinks=true,
    linkcolor=black,
    citecolor=black,
    filecolor=cyan,
    urlcolor=blue
}

\usepackage{tcolorbox}

\title{\LARGE \bf
Symmetry Considerations for Learning Task Symmetric Robot Policies
}

\author{Mayank~Mittal$^*$, Nikita~Rudin$^*$, Victor~Klemm, Arthur~Allshire, and~Marco~Hutter%
\thanks{
$^*$~M. Mittal and N. Rudin contributed equally.
} %
\thanks{
This work was supported by the Swiss National Science Foundation through the National Centre of Competence in Digital Fabrication (NCCR dfab). It has also received funding from the European Research Council (ERC) under the European Union’s Horizon 2020 research and innovation programme grant agreement No. 852044.
}%
\thanks{All authors are with the Robotic Systems Lab, ETH Z\"urich, 8092 Z\"urich, Switzerland.  A. Allshire is with the University of Toronto, Canada. M. Mittal, N. Rudin, and A. Allshire are also with NVIDIA.}
\thanks{Contact: {\tt\footnotesize \{mittalma, rudinn, vklemm\}@ethz.ch}}
}

\begin{document}

\maketitle
\thispagestyle{empty}
\pagestyle{empty}

\begin{abstract}

Symmetry is a fundamental aspect of many real-world robotic tasks. However, current deep reinforcement learning (DRL) approaches can seldom harness and exploit symmetry effectively. Often, the learned behaviors fail to achieve the desired transformation invariances and suffer from motion artifacts. 
For instance, a quadruped may exhibit different gaits when commanded to move forward or backward, even though it is symmetrical about its torso.
This issue becomes further pronounced in high-dimensional or complex environments, where DRL methods are prone to local optima and fail to explore regions of the state space equally.
Past methods on encouraging symmetry for robotic tasks have studied this topic mainly in a single-task setting, where symmetry usually refers to symmetry in the motion, such as the gait patterns.
In this paper, we revisit this topic for goal-conditioned tasks in robotics, where symmetry lies mainly in task execution and not necessarily in the learned motions themselves.
In particular, we investigate two approaches to incorporate symmetry invariance into DRL -– data augmentation and mirror loss function.
We provide a theoretical foundation for using augmented samples in an on-policy setting. Based on this, we show that the corresponding approach achieves faster convergence and improves the learned behaviors in various challenging robotic tasks, from climbing boxes with a quadruped to dexterous manipulation.

\end{abstract}

\section{Introduction}

Deep reinforcement learning (DRL) is becoming an important tool in robotic control.
Without prior knowledge or any assumptions on the underlying model, these methods can solve complex tasks such as legged locomotion \cite{miki2022learning,LeeLearningQuadrupedal,kumar2021rma}, object manipulation \cite{allshire2021transferring,akkaya2019solving}, and goal navigation \cite{rudin2022advanced}.
However, this very black-box nature of DRL does not leverage the knowledge of the symmetry in the task and often results in policies that are not invariant under symmetry transformations~\cite{YuLearningSymmetric,vanderpool2020homonet}.
This problem is not limited to the current DRL algorithms. Humans and animals also exhibit asymmetric execution of various tasks by, for example, always using the dominant hand or foot for tasks requiring higher dexterity. Robots, however, should avoid such limitations and achieve optimal task execution in all cases.

\begin{figure}
    \centering
    \includegraphics[width=\linewidth]{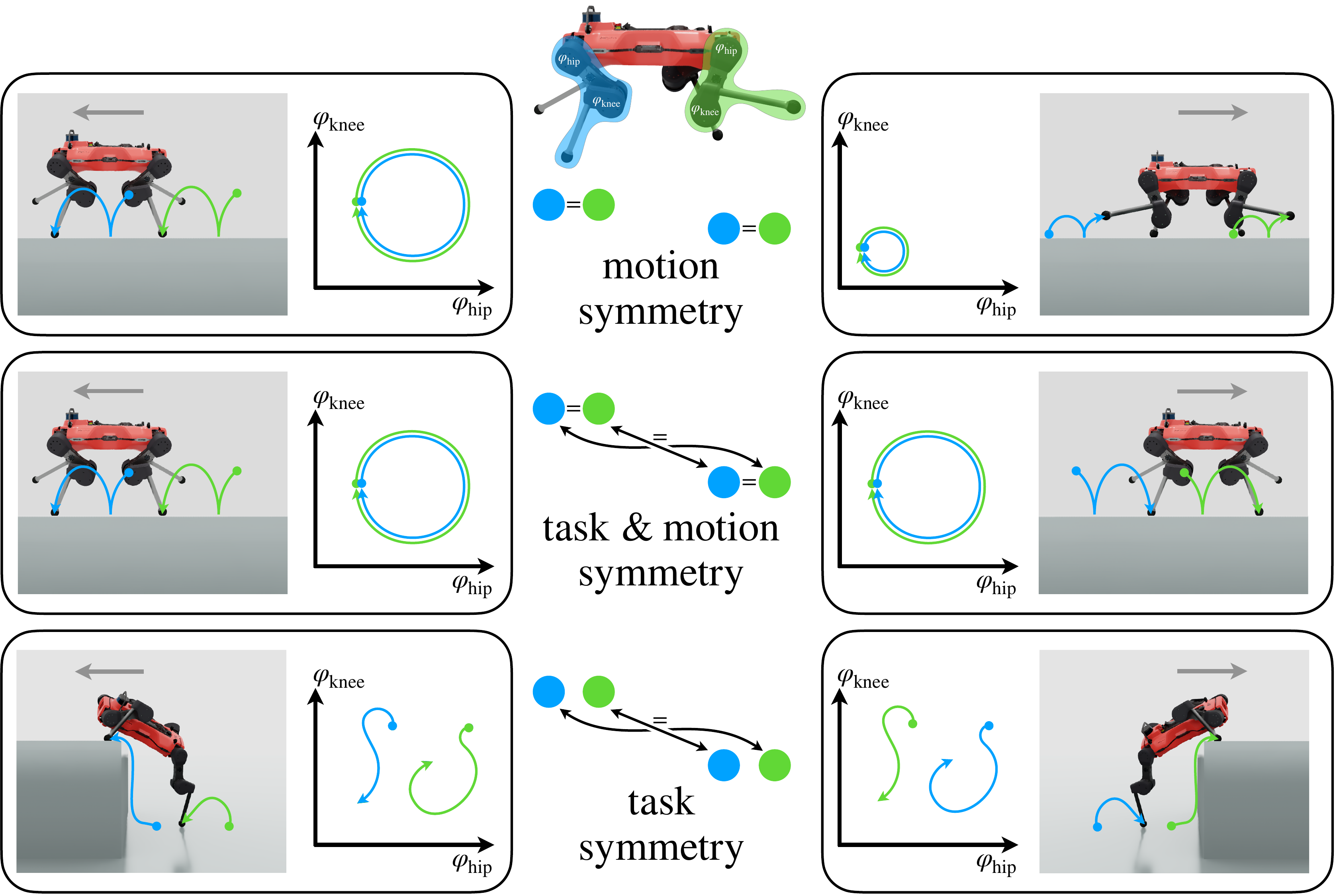}
    \caption{Motion and task symmetry for quadrupeds. While motion symmetry involves similar movements of the legs, it does not guarantee that the robot behaves the same when commanded different goals (walking forward and backward). In contrast, task symmetry ensures consistent behaviors for such goals, potentially resulting in periodic symmetric motions for walking on flat ground or entirely asymmetrical aperiodic patterns for tasks such as climbing a box.}

    \label{fig:intro-figure}
\end{figure}

In robotics, we can think of symmetry at two levels: 1) motion execution, which pertains to the behavior of mirrored body parts during periodic motions, and 2) task execution, which pertains to the behavior used to achieve mirrored objectives. This distinction is crucial since achieving symmetry in task execution does not necessarily imply or demand symmetry in motion execution. To illustrate, consider tasks for quadrupedal locomotion (\figref{fig:intro-figure}).
A typical locomotion task may display both symmetries by learning a trotting gait for all commanded directions~\cite{LeeLearningQuadrupedal}. However, when faced with the challenge of climbing a tall box, the robot needs to deviate from symmetry at the motion level~\cite{rudin2022advanced}. Nevertheless, it can still maintain symmetry at the task level; for instance, climbing a box in front of or behind the robot is considered equivalent.
While we anticipate that behaviors for symmetrical goals will exhibit similarities, the solutions obtained using DRL are not. Usually, the trained policies exploit the behavior learned for only one of the goals. For instance, instead of climbing the box backward, the robot may first turn around and then ascent the box. Unfortunately, this behavior consumes more time and energy, rendering it sub-optimal. During the learning process, once an asymmetry in a behavior arises, it tends to get magnified with further training. Hence, it is important to incorporate symmetry considerations inherent to the task into DRL to learn superior and more efficient behaviors.

\subsection{Related Work}
\label{sec:related_work}

Achieving symmetric motions has been of long-standing interest in character animation and, recently, robotics, where symmetrical gaits are usually considered more visually appealing and efficient. In model-based control, symmetric motions are typically enforced by hard-coding gaits~\cite{CorosLocomotionSkills} or by reducing the optimization problem by assuming perfect symmetry~\cite{MajkowskaFlipping}.
Similarly, in robot learning, the structure of the action space can be modified to ensure a symmetric policy. For instance, central pattern generation (CPG) for locomotion pushes the policy towards symmetrical sinusoidal motions~\cite{LeeLearningQuadrupedal, BellegardaCPG}. 
For periodic motions, motion phases as a function of time can also be used to learn policies for only half-cycles and repeat them during execution~\cite{abdolhosseini2019learning, liu2016guidedlearning}.
Alternatively, based on the robot's morphology, the policy can control only half of the robot, with the other half simply repeating the selected actions~\cite{Rudin2021CatLike}. While these ideas are simple, they constrain the policy by some explicit switching mechanism based on time or behavioral patterns.

To avoid this issue, recent works have looked at introducing invariance to symmetry transformations into the learning algorithm itself. Inspired by the success of data augmentation in deep learning, one way to induce this invariance is by augmenting the collected experiences with their symmetrical copies~\cite{abdolhosseini2019learning,lin2020ker}. An alternate approach is adding a penalty or loss function to the learning objective~\cite{YuLearningSymmetric,abreu2023addressing}. It is also possible to design special network layers to represent functions with the desired invariance properties~\cite{vanderpool2020homonet,wang2021incorporating,wang2022so2equivariant,ordonez2023discrete}. Abdolhosseini~\etal~\cite{abdolhosseini2019learning} compared these different approaches for bipedal walking characters. They showed that in many cases, using a symmetry loss function is more effective than data augmentation and performs at par with customized network architectures.

It is important to note that most of the above works have studied symmetry under the lens of symmetrical motions, or more specifically, gait patterns. This may not always be desired or feasible for a wider range of tasks, such as manipulating objects or climbing over surfaces, where symmetry appears at the task level and not on how symmetrically located actuators move. This paper aims to revisit the idea of symmetry from this task perspective and understand its efficacy on different real-world robotic problems.

\subsection{Contributions}

We investigate the notions of symmetry in DRL for goal-conditioned tasks.
Specifically, we explore two approaches for embedding symmetry invariance into on-policy RL: data augmentation and mirror loss function.
While these methods have previously appeared in literature, their applications have primarily centered around walking animated characters, rather than robotic tasks with goal-level symmetries.
Our analysis aims to highlight often-overlooked intricacies in the implementations of these approaches.
In particular, we discuss the ineffectiveness of naive data augmentation and introduce an alternate update rule that helps stabilize learning from augmented samples.

Our study compares the two approaches on four diverse robotic tasks: the standard cartpole, agile locomotion with a quadruped, object manipulation with a quadruped, and dexterous in-hand object manipulation.
Notably, in contrast to prior work~\cite{abdolhosseini2019learning}, our experimental findings show that data augmentation is the most effective way to achieve task-symmetrical policies.
We demonstrate the sim-to-real transfer of policies learned with this method for agile locomotion using the platform ANYmal~\cite{hutter2017anymal}. Although the robot is not perfectly symmetrical, we show that the policy trained using data augmentation results in nearly symmetrical behaviors for climbing boxes in front of and behind the robot.

\section{Preliminaries}
\label{sec:preliminaries}

\subsection{Reinforcement Learning}
\label{sec:rl-prelims}

This work considers robotic tasks modeled as multi-goal Markov Decision Processes (MDPs) with continuous state and action spaces. For notational simplicity, we consider the goal specification a part of the state definition. We denote an MDP $\mdp$ as $(\mathcal{S}, \mathcal{A}, \T, \r, \gamma, \rho_0)$, where the symbols follow their standard definitions~\cite{sutton2018rl}.
Our goal is to obtain a policy $\pi$ that maximizes the expected discounted reward, $J(\pi) = \Eb{\tau \sim p_{\pi}(\tau)}{\sum_{t=0}^{\infty} \gamma^t r(s_t, a_t)}$, where the trajectory $\tau = (s_0, a_0, s_1, a_1, s_2, \dots)$ is sampled from $p_{\pi}(\tau)$ with $s_0 \sim \rho_0(\cdot),\ a_{t} \sim \pi(\cdot| s_t),$ and $s_{t+1} \sim T(\cdot |  s_t, a_t)$.
As before, we employ the definitions from~\cite{sutton2018rl} for the state-action value function $\Qpi(s_t, a_t) = \Eb{s_{t+1}, a_{t+1},\dots}{\sum_{l=0}^{\infty} \gamma^l r(s_{t+l}, a_{t+l})}$, the value function $\Vpi(s_t) = \Eb{a_{t}}{\Qpi(s_t, a_t)}$, and the advantage function $\Api(s_t, a_t) = \Api_t = \Qpi(s_t, a_t) - \Vpi(s_t)$.

In DRL, the total expected reward can only be estimated through trajectories collected by executing the current policy $\pithetk$, where $\theta_k$ are the policy's parameters at the learning iteration $k$. Following this, modern policy gradient approaches, such as TRPO~\cite{schulman2015trpo} and PPO~\cite{schulman2017proximal}, use importance sampling to rewrite the policy gradient as:
\begin{align}
    \label{equ:policy-grad}
    & \gradth J(\pithet) = \Eb{\tau \sim p_\pithetk}{\sum_{t=0}^{\infty} \eta_{t}(\theta) \Apithetk_t \gradth \log{\pithet}(a_t|s_t)},  \nonumber \\
    & \text{where } \eta_{t}(\theta) = \frac{p_\pithet(s_t, a_t)}{p_\pithetk(s_t, a_t)} = \frac{p_\pithet(s_t)}{p_\pithetk(s_t)}  \frac{\pithet( a_t | s_t)}{\pithetk( a_t | s_t)}.
\end{align}
In practice, the term $\frac{p_\pithet(s_t)}{p_\pithetk(s_t)}$ is computationally intractable. However, it can be neglected by assuming the divergence between the policy distributions $\pithet$ and $\pithetk$ is sufficiently small~\cite{schulman2015trpo}. In PPO, this is achieved by using a clipped surrogate loss, $\loss^{\text{PPO}}(\theta)$~\cite{schulman2017proximal}.
Additionally, the value function is fitted using a supervised learning loss.

\subsection{MDP with Group Symmetries}

For an MDP $\mdp$ with symmetries, a set of transformations exists on the state-action space, such that the reward function and transition dynamics are invariant to them~\cite{ravindran2001symm, zinkevich2001symm}. More formally, we define a symmetric MDP with an N-fold symmetry if it contains a set of symmetric transformations $\mathcal{G} = \cup_{k} G_k = \{g_0, g_1, g_2, \dots g_{N-1}\}$, where $g_0:= (\mathbb{I}, \mathbb{I})$ is the identity transformation, and $g_i:= (L_{g_i}, K_{g_i}), \forall i \in \{1, \dots, N-1\}$ are distinct non-identity transformations. The operators $L_g: \S \rightarrow \S$ and $K_g: \A \rightarrow \A$ can be seen to define similar transformations but in different spaces.

\section{Approaches for Symmetry in RL}
\label{sec:method}

In literature, there are three main ways to incorporate symmetry into DRL: 1) using a symmetry loss function, 2) performing data augmentation, and 3) designing specialized network architectures. While the first two approaches only approximate the symmetry equivariance, specialized networks tend to guarantee it by embedding the equivariances into the layers themselves. However, this constrains the policy to always be equivariant, which can be detrimental in robotic applications since robots are not perfectly symmetrical. Additionally, perfectly symmetrical policies struggle with neural states, where $s = L_g[s], \forall g \in \mathcal{G}$, unless the environment introduces its own bias~\cite{yan2023geometric}. For instance, consider a quadruped starting to walk from a stance gait. A symmetric policy cannot lift the right front foot to take the first step since that means the other feet should also be raised under ${L_g[s]}_{g \in \mathcal{G}}$. However, this is not possible since $\pi(s) \neq \pi(L_g[s]), \forall g \in G - \{g_0\}$.

In practice, we only want to encourage the policy to learn similar behaviors for equivalent goals while letting it adapt the individual actuation or motion-level commands to deal with the asymmetries in the robot's design and neutral states. Keeping this in mind, we mainly look at the symmetry loss function and data augmentation approaches.

\subsection{Using Mirror Loss Function}

In the method proposed by Yu~\etal~\cite{YuLearningSymmetric}, they add an explicit auxiliary loss to the learning objective that penalizes asymmetricity in the policy. Based on this approach, we can write the policy learning objective for all symmetry transformations in $\mathcal{G}$ as:
\begin{align}
    & \loss(\theta) = \loss^{\text{PPO}}(\theta) + w \sum_{g \in \mathcal{G}} \loss^{\text{sym}}_g(\theta), \text{where} \label{equ:symm-loss-obj} \\
    & \loss^{\text{sym}}_g(\theta) = \Eb{\tau \sim p_\pithetk}{\sum_{t=0}^\infty \normtwo{K_g [\pithet(s_t)] - \pithet( L_g[s_t])}^2}, \label{equ:symm-loss}
\end{align}
and $w$ is a scalar hyperparameter that governs the trade-off between minimizing the RL objective and the symmetry loss. Tuning this parameter $w$ depends on the task and can adversely affect the training if set to a high value. Although not explicitly mentioned in prior works~\cite{YuLearningSymmetric}, during implementation, the quantity $K_g[\pithet(s)]$ is treated as a label and is not back-propagated through, despite its differentiability.

From an intuitive standpoint, the symmetry loss (\eqtref{equ:symm-loss}) encourages the policy to be symmetrical over its entire state-action spaces. However, achieving this objective can be challenging in high-dimensional problem spaces.

\subsection{Symmetry-Based Data Augmentation}
\label{sec:sym-aug-method}

Data augmentation is commonly used in deep learning to make networks invariant to visual or geometrical transformations~\cite{krizhevsky2012imagenet,laskin2020rad}.
A natural approach for symmetry augmentation within RL is augmenting the collected trajectories with their symmetrical copies~\cite{abdolhosseini2019learning}.
However, this results in having to evaluate $\pithetk(\cdot | \cdot)$ and $\Apithetk(\cdot, \cdot)$ in~\eqtref{equ:policy-grad} on samples not generated from the rollout policy.
Computing these quantities using such ``off-policy'' samples can introduce high variance in the gradients and diminish the method's effectiveness~\cite{abdolhosseini2019learning}.

\begin{figure}
  \centering
  \begin{minipage}[c]{0.45\linewidth}
    \includegraphics[width=1.1\textwidth, trim=8 12 8 12, clip]{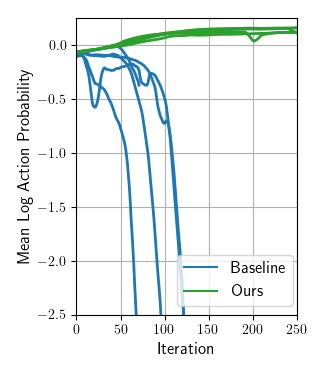}
  \end{minipage}
  \quad
  \begin{minipage}[c]{0.475\linewidth}
      \caption{The log action probabilities computed using baseline~(\eqtref{equ:policy-grad}) and our proposed (\eqtref{equ:sym-aug}) approaches. We plot the mean obtained over the symmetry-augmented samples from each training iteration. The plot shows 5 runs with different seeds for the CartPole task. The baseline method leads to training instabilities caused by low action probabilities. Meanwhile, our approach maintains stable convergence for all runs.}
    \label{fig:data-augmentation}
  \end{minipage}
\end{figure}

To deal with this issue, we approach symmetry augmentation from another perspective. At iteration $k$, let us construct policies $\pithetkg$, such that $\pithetkg(K_g[a] | L_g[s]) = \pithetk(a | s), \forall g \in \mathcal{G}, \s \in \S, \a \in \A$. Based on these augmented policies, we can write the RL objective for $\pithet$ (\eqtref{equ:policy-grad}) as learning from trajectories collected from these policies, \ie $\tau^g = (s_0^g, a_0^g, \dots)$:
\begin{align}
    \label{equ:multi-pi-g}
    & \gradth J(\pithet) = \sum_{g \in \mathcal{G}} \Eb{\tau^g \sim p_\pithetkg}{\sum_{t=0}^{\infty} \eta_{t}^g(\theta) \Apithetkg_t \gradth \log{\pithet}(a^g_t|s^g_t)},  \nonumber \\
    & \text{where } \eta_{t}^g(\theta) = \frac{p_\pithet(s_t^g, a_t^g)}{p_\pithetkg(s_t^g, a_t^g)} = \frac{p_\pithet(s_t^g)}{p_\pithetkg(s_t^g)}  \frac{\pithet( a_t^g | s_t^g)}{\pithetkg( a_t^g | s_t^g)}.
\end{align}
In data augmentation, the samples are collected by rolling out $\pithetk$ and not $\pithetkg$, \ie $\tau^g = (s_0^g, a_0^g, \dots) = (L_g[s_0], K_g[a_0], \dots)$ with $s_t, a_t \sim p_\pithetk(s_t, a_t), \forall t \geq 0$.

Additionally, for the policies $\pithetkg$ and the symmetric MDP $\mdp$, it can be shown that $\forall s \in \S, a \in A, g \in \mathcal{G}$:
\begin{align}
    \label{equ:pi-g-mdp}
    & A^{\pithetkg}(L_g[s], K_g[a]) = A^{\pithet}(s, a) \neq A^{\pithet}(L_g[s], K_g[a]), \text{and } \nonumber \\
    & p_{\pithetkg}(L_g[s]) = p_\pithetk(s) \neq p_\pithetk(L_g[s]).
\end{align}
Thus, using \eqtref{equ:multi-pi-g} and \eqtref{equ:pi-g-mdp}, we obtain:
\begin{align}
    \label{equ:sym-aug}
    &\gradth J(\pithet) = \sum_{g \in \mathcal{G}} \mathbb{E}_{\tau \sim p_\pithetk}\Biggl[\sum_{t=0}^{\infty}
    \frac{p_\pithet(L_g[s_t])}{p_\pithetk(s_t)}  
    \frac{\pithet( K_g[a_t] | L_g[s_t])}{\pithetk( a_t | s_t)} \nonumber \\
    & \qquad \quad \Apithetk(s_t, a_t) \gradth \log{\pithet}(K_g[a_t] | L_g[s_t])\Biggr].
\end{align}

Comparing \eqtref{equ:sym-aug} to simply applying \eqtref{equ:policy-grad} on augmented samples, we can see that the denominator of the action probability ratio are different. Using \eqtref{equ:policy-grad}, we would get $\frac{\pithet( K_g[a_t] | L_g[s_t])}{\pithetk( K_g[a_t] | L_g[s_t])}$, while with \eqtref{equ:sym-aug}, we have $\frac{\pithet( K_g[a_t] | L_g[s_t])}{\pithetk( a_t | s_t)}$. In other words, \eqtref{equ:sym-aug} keeps the action probability of the original samples. In contrast, we would need to compute the action probability for augmented samples for the other case. This difference is crucial since $\pithetk( K_g[a_t] | L_g[s_t])$ can be arbitrarily small for not perfectly symmetric policies, leading to instabilities in the training, as shown in \figref{fig:data-augmentation}.

However, even with the above change, the issue with computing the probability ratio $\frac{p_\pithet(L_g[s_t])}{p_\pithetk(s_t)}$ remains unresolved. It can only be disregarded if the constructed policies $\{\pithetkg\}_{g \in \mathcal{G}}$ are sufficiently close to the policy $\pithetkg$ used for generating rollouts. While this may not hold for any policy $\pithetk$, from our experiments in~\secref{sec:net-weights}, we find that the probability ratio term can be ignored in the case of randomly initialized policies with sufficiently small weights and bounded updates. However, the ratio is important when policies are initialized non-symmetrically.

Conceptually, we can interpret \eqtref{equ:sym-aug} as follows: When we observe a high return for a specific action $a$ taken from a given state $s$, we want to boost the likelihood of choosing that action in the future. In the case of symmetry, we also want to amplify the likelihood of the equivalent action $K_g[a]$ taken from the equivalent state $L_g[s]$.

\section{Experiments and results}
\label{sec:exp}

\subsection{Tasks}
\label{sec:exp-setup}

\begin{figure}
    \includegraphics[width=0.2425\linewidth,trim={0cm 2.25cm 0 2cm},clip]{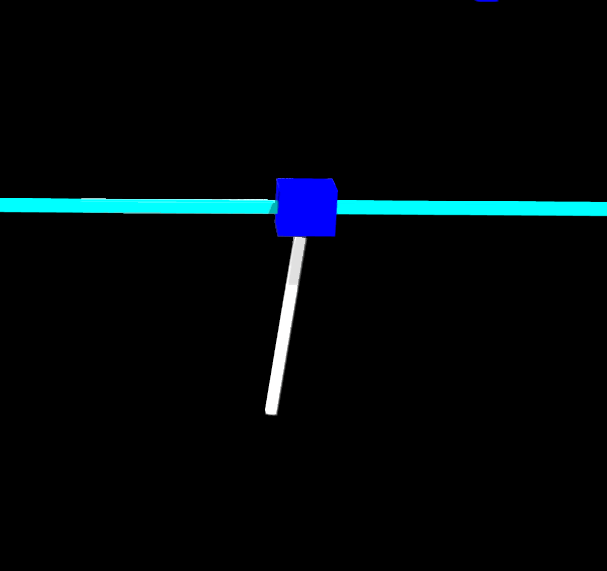}
    \includegraphics[width=0.2425\linewidth]{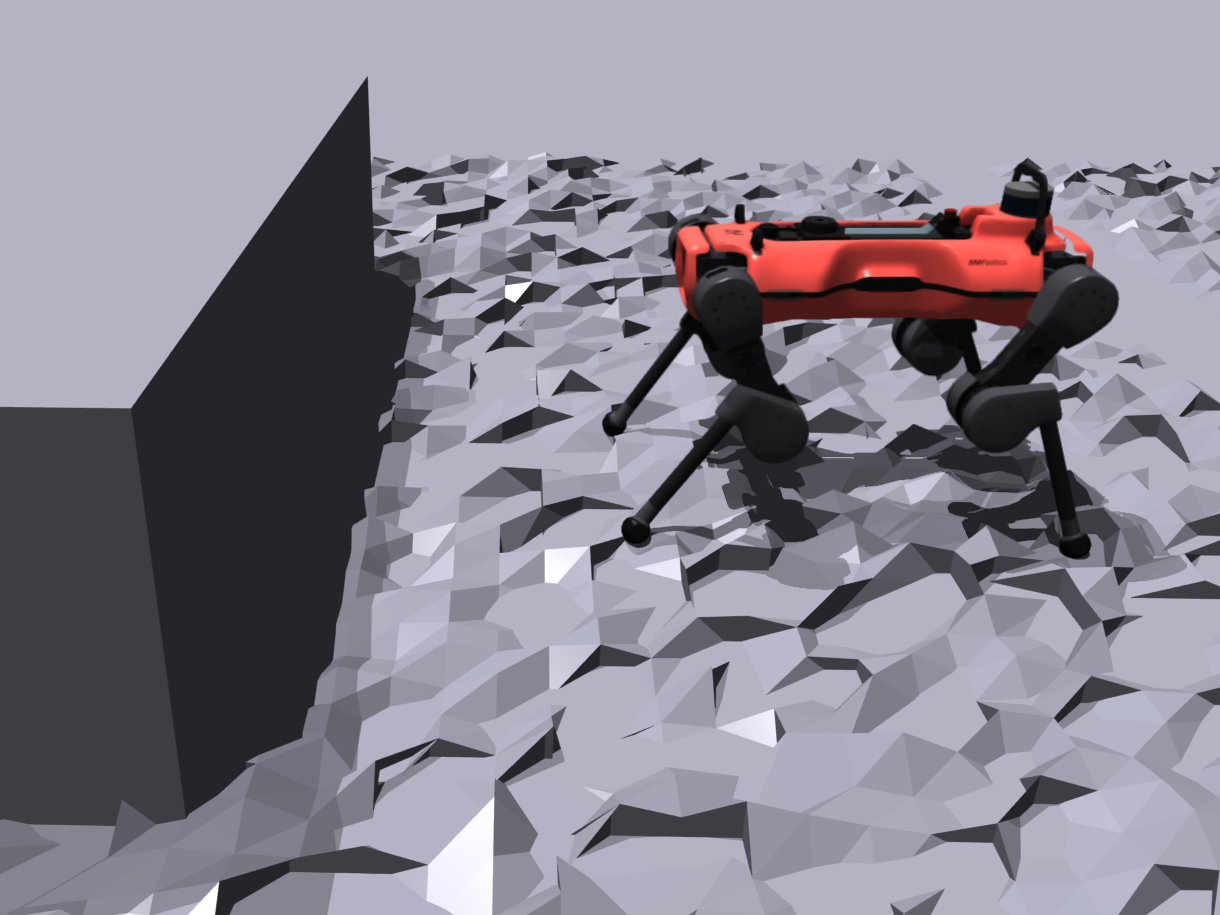}
    \includegraphics[width=0.2425\linewidth,trim={10cm 5.5cm 0 2cm},clip]{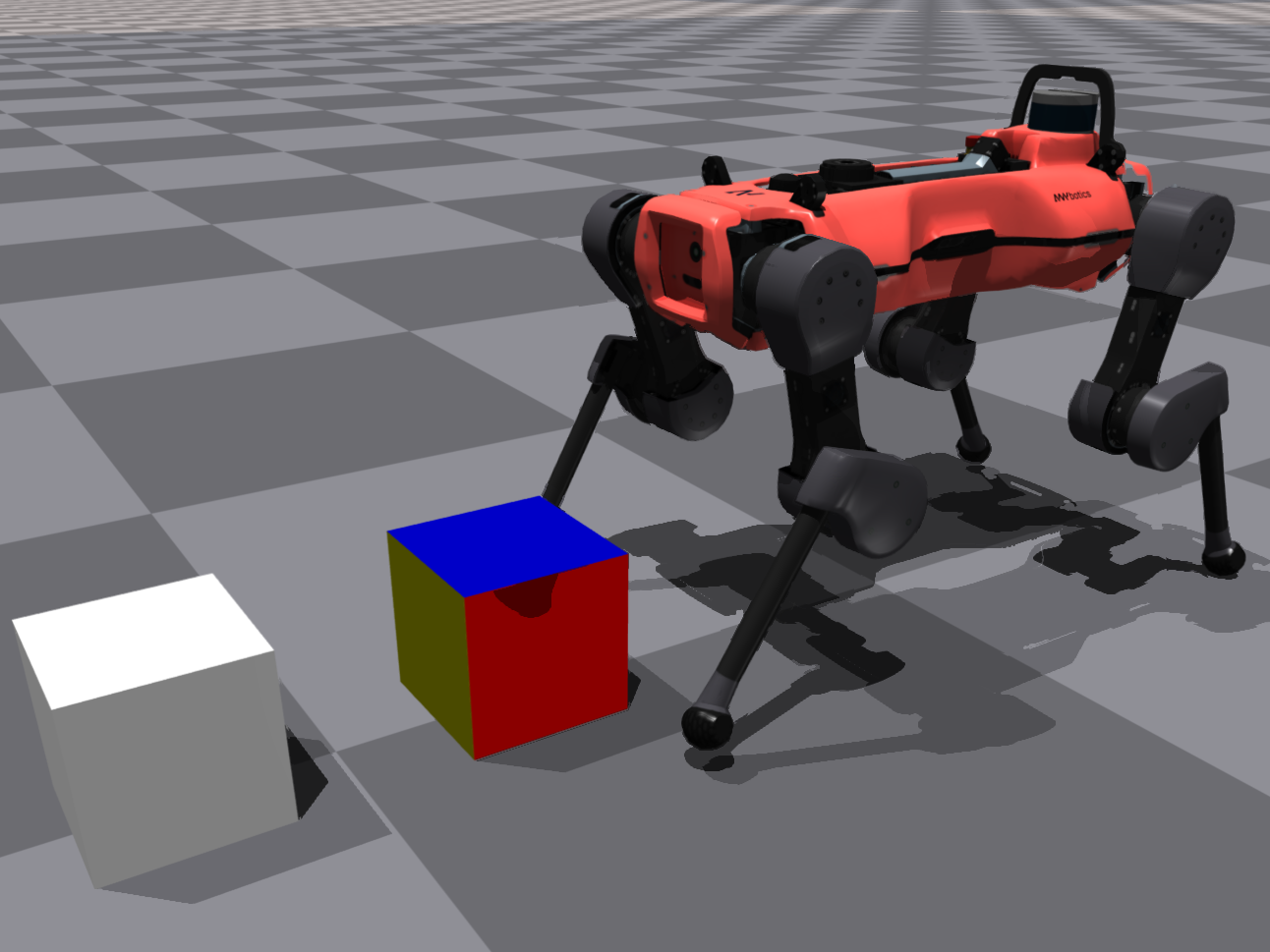}
    \includegraphics[width=0.2425\linewidth]{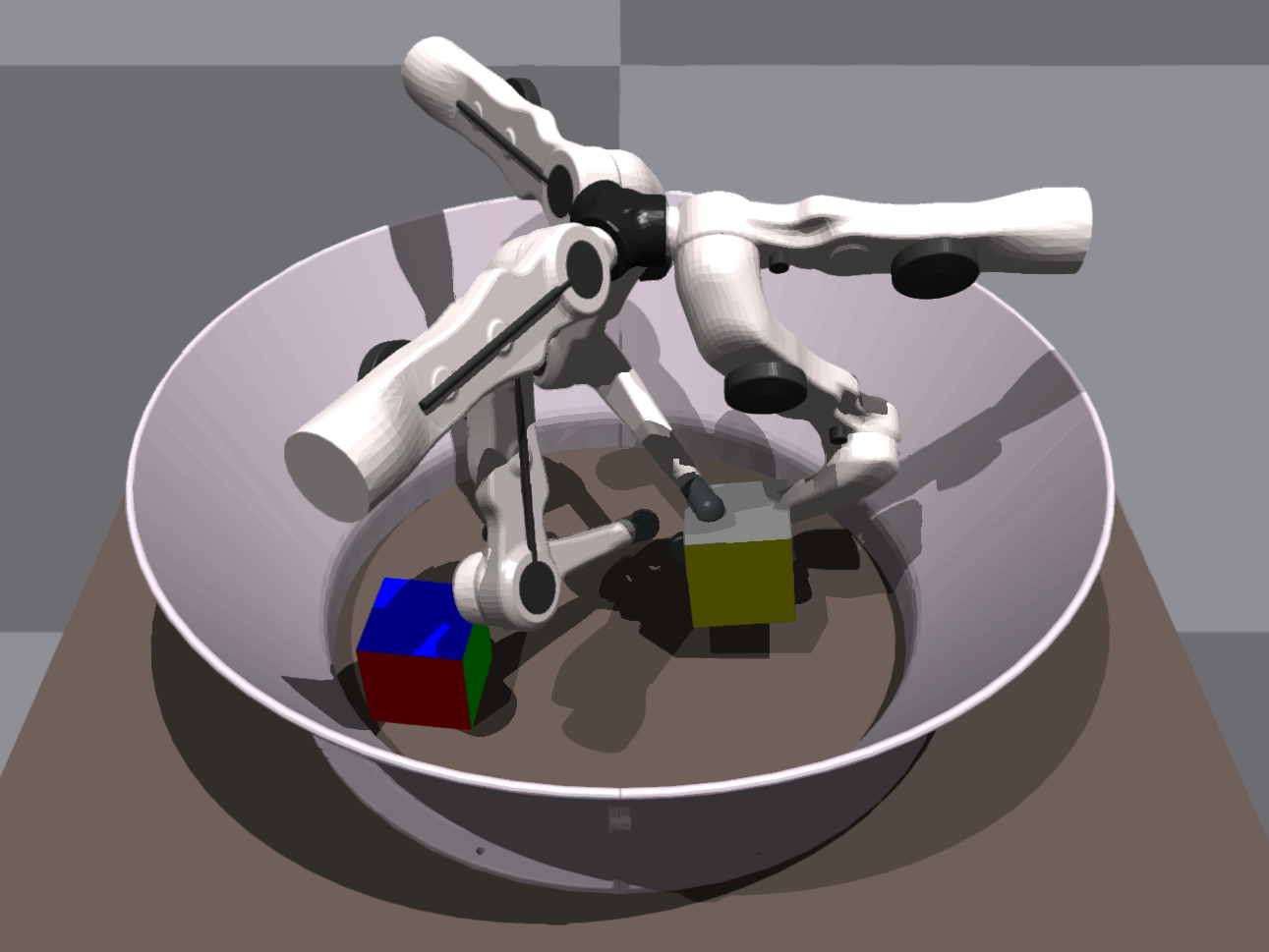}
    \resizebox{\columnwidth}{!}{%
    \begin{tabular}{rlll}
    \toprule
        Task  && Space  & Transformations \\
    \midrule
        CartPole & $\S$ &$(\dot{x},\theta,\dot{\theta})$    & $(\dot{x},\theta,\dot{\theta}),(-\dot{x},-\theta,-\dot{\theta})$ \\
        & $\A$ & $(\dot{x}_\des)$      & $(\dot{x}_\des), (-\dot{x}_\des)$ \\
        \rowcolor[HTML]{EFEFEF} ANYmal-Climb & $\S$ & $\Rsp^{282}$    & Identity, \quad reflect-x, \quad reflect-y, \quad $\curvearrowright 180^\circ$ \\
        \rowcolor[HTML]{EFEFEF} & $\A$ & $\Rsp^{12}$      & Identity, \quad reflect-x, \quad reflect-y, \quad $\curvearrowright 180^\circ$ \\
        ANYmal-Push & $\S$ & $\Rsp^{51}$   & Identity, \quad reflect-x, \quad reflect-y, \quad $\curvearrowright 180^\circ$ \\
        & $\A$ & $\Rsp^{12}$ & Identity, \quad reflect-x, \quad reflect-y, \quad $\curvearrowright 180^\circ$ \\
        \rowcolor[HTML]{EFEFEF} Trifinger-Repose& $\S$ & $\Rsp^{41}$ & Identity, \quad $\curvearrowright 120^\circ$, \quad$\curvearrowright 240^\circ$ \\
        \rowcolor[HTML]{EFEFEF} & $\A$ & $\Rsp^{9}$ & Identity, \quad $\curvearrowright 120^\circ$, \quad$\curvearrowright 240^\circ$  \\
    \bottomrule
    \end{tabular}
    }
    \caption{We consider four robotic tasks: a continuous cart-pole, quadruped climbing a box, quadruped manipulating a cube, and in-hand cube reposing. In the table, we specify their state and action spaces along with the available symmetry transformations.}
    \label{fig:all-tasks}
\end{figure}

We consider four tasks, implemented using NVIDIA Isaac Gym ~\cite{makoviychuk2021isaac}, with inherent task symmetry (\figref{fig:all-tasks}):

\begin{itemize}
    \item \textit{CartPole}: A classic control environment where the goal is to balance a pole attached by an unactuated joint to a cart. The input to the system is the desired cart velocity. As a reward, the agent receives an L-1 penalty between the pole's current and upright position. 
    \item \textit{ANYmal-Climb}: An agile quadrupedal locomotion task from~\cite{rudin2022advanced}, where the quadruped ANYmal~\cite{hutter2017anymal} needs to reach a target pose on a box over a defined time. The agent observes its state along with a robot-centric height map and receives a sparse delayed reward signal.
    \item \textit{ANYmal-Push}: A loco-manipulation task where the robot needs to push a cube to a desired position. The cube's initial and target positions are spawned radially around the robot. The agent observes the robot's and object's states and receives a dense tracking reward.
    \item \textit{Trifinger-Repose}: An in-hand cube reposing task for the Trifinger platform~\cite{wuthrich2021trifinger}. The agent needs to pick the cube from the table and manipulate it to its desired pose. The task setup is similar to that in~\cite{allshire2021transferring}.
\end{itemize}

The two quadrupedal tasks use curriculums to guide the training. For the climbing task, we use an initial move-in-direction reward that encourages the robot to move toward the target pose (phase A). This reward is later removed so that the robot can optimize its motion freely (phase B), as done in \cite{rudin2022advanced}. Once the robot starts climbing the box successfully, we randomize its initial orientation (phase C). Instead of always facing the boxes (yaw $=0$), the orientation is sampled uniformly with yaw $\in[-\pi, \pi]$. Note that this curriculum intervention is necessary to achieve effective climbing behaviors. When training policies with randomized orientations from the beginning, they converge to a sub-optimal sideways climbing motion, which fails to solve the task for higher boxes.
For the \textit{ANYmal-Push} task, the curriculum moves the cube target further away as the robot pushes the cube successfully.

\subsection{Metrics}

Prior work~\cite{abdolhosseini2019learning} uses metrics that typically characterize the gait symmetricity. However, this does not serve as a proper measure for a policy's symmetricity during task execution. For instance, in the \textit{ANYmal-climb} task, we do not require that the front and back legs follow similar trajectories, but rather that when climbing forward and backward, the front legs behave similarly to the back legs, respectively.

Thus, we use two metrics that directly characterize the policy's performance in the task and measure its symmetricity:
1) the \textit{average episodic return}, which is the undiscounted reward accumulated by the policy over an episode, and 2) the \textit{symmetry loss} from \eqtref{equ:symm-loss}, which measures the discrepancy in the policy for equivalent state-action pairs.

\begin{figure*}[t]
    \centering
    \includegraphics[width=0.935\linewidth]{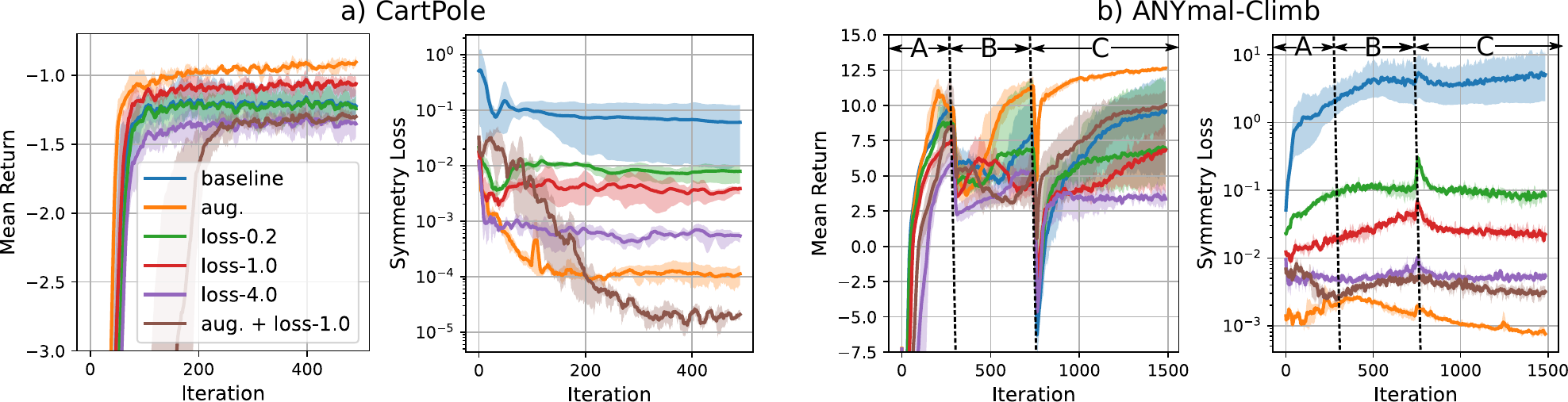}
    \caption{Comparison of different methods for the CartPole and ANYmal-Climb tasks -- vanilla PPO (baseline), PPO with symmetry augmentation (aug.), PPO with symmetry loss (\textsf{loss-w}), and a combination of the two. We plot the mean and standard deviation over three seeds. For the ANYmal-Climb task, we use a curriculum denoted as phases A, B, and C in the plot. We observe that symmetry augmentation yields the best performance consistently over all the tasks.}
    \label{fig:training-curves}
    \vspace{-4pt}
\end{figure*}

\subsection{Training Performance}

We compare PPO with symmetry loss, symmetry augmentation, and a combination of both against the standard version of the algorithm~\cite{schulman2017proximal}. For PPO with symmetry loss, we consider different weights $w$ to understand its implications. We use the weight from the best policy for the combined symmetry augmentation and loss method.

From \figref{fig:training-curves},  we observe that PPO with symmetry augmentation obtains the highest return and fastest convergence while having a low symmetry loss. Optimizing the symmetry loss directly helps induce symmetry but comes at the cost of performance and slower convergence. Increasing the weight $w$ reduces the symmetry loss but hinders learning as the gradients from the losses in \eqtref{equ:symm-loss-obj} compete against each other. 

Additionally, for the ANYmal-Climb task, we can notice how different methods recover once phase C begins. At the start of this phase, the sudden change in the robot's orientation causes all the policies to fail since they now need to perform the climbing motion in different directions. The policy training with symmetry augmentation recovers nearly immediately as it is inherently symmetric from being trained on other orientations through the augmented samples. It must only adapt to intermediate orientations not previously seen in the earlier phases. On the other hand, policy training with the vanilla PPO takes much longer to recover and converges to a different behavior (\secref{sec:qualitative-behavior}). Lastly, the policies trained using symmetry loss do not consistently recover in this phase. %

Notably, the symmetry loss weighs symmetricity equally for all equivalent state actions. During training, the policy explores new actions for each symmetric state independently. If better actions are found for one of the states, the symmetry loss will push the policy to adopt equivalent actions for all equivalent states without considering the respective rewards. On the other hand, the augmentation approach will push the policy towards the best-performing actions since all transitions are compared to the same value function. Interestingly, using both symmetry loss and augmentation does not necessarily improve the performance or convergence, showing that symmetry augmentation does not benefit from the additional gradients provided by the loss.

\subsection{Effect of network initialization}
\label{sec:net-weights}

As discussed in~\secref{sec:sym-aug-method}, symmetry augmentation assumes that the rolled-out policy is sufficiently symmetric, and hence, the slightly off-policy samples do not cause issues during training. A symmetric policy is expected to maintain that characteristic throughout training. However, when training commences from an arbitrary policy, there is no guarantee that it will converge to exhibit symmetric behaviors. To assess the severity of this problem, we compare the training of policies initialized with randomized weights drawn from a uniform distribution with varying scales.

For small weights, the actions from the policy are typically small as well, and as such, the policy is roughly equivalent to its symmetric counterparts. More concretely, for Gaussian distributions, policies $\pi_\theta(a|s)$ and $\pi_\theta(L_g[a] | K_g[s])$ are similar for small means and large enough standard deviation. With larger weights, the disparity between the two distributions increases, and they diverge from each other. 

\begin{figure}[t]
    \centering
    \includegraphics[width=0.98\linewidth]{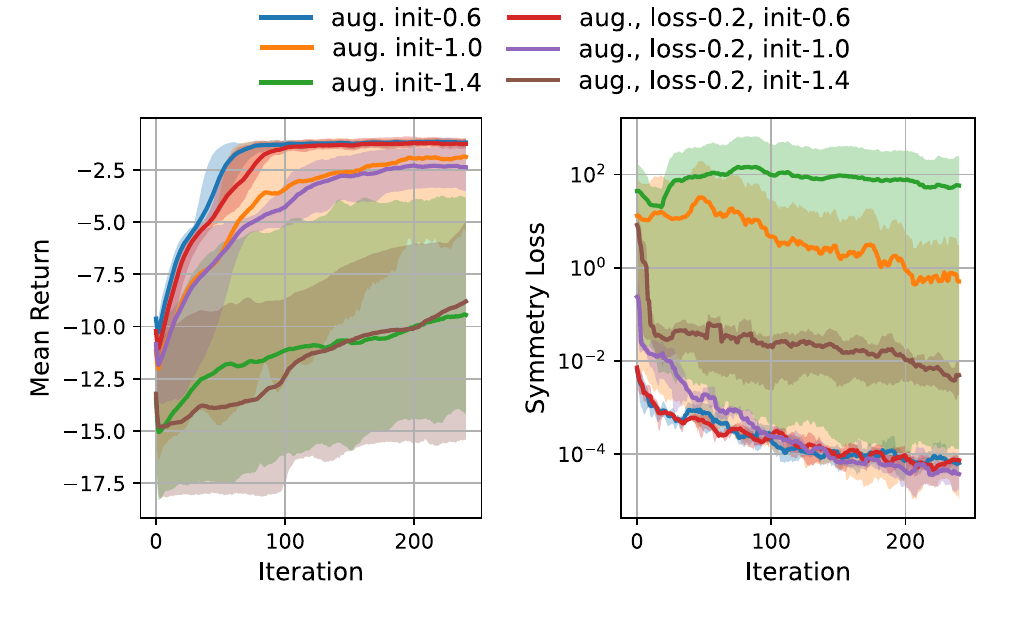}
    \caption{Effect of network initialization scales (\textsf{init-n}) for the CartPole task. We plot the mean and standard deviation over three seeds. Symmetry augmentation (aug.) struggles when initialized weights are high. Adding a small symmetry loss helps mitigate the issue but does not improve the performance.}
    \label{fig:network-init}
\end{figure}

\figref{fig:network-init} shows that, indeed, the scale of initial weights influences the performance and symmetricity of the policies trained with data augmentation. Higher weights lead to lower performance and higher symmetry loss.
Since directly optimizing over the symmetry loss does not assume any symmetricity of policy, we speculate that it is not affected by the initialization effect, and combining both augmentation and loss approaches can help recover the symmetricity even when the policy is initialized with high weights. Our findings, shown in \figref{fig:network-init}, affirm that using a small symmetry loss coefficient greatly enhances the symmetricity of policies initialized with high weights.
However, it is worth noting that this enhancement does not translate into improved performance compared to policies with low-weight initializations.

\subsection{Evaluation of Symmetry in Learned Behaviors}

To evaluate the performance of policies trained with and without symmetry augmentation, we create equivalent versions of each task and compute their total episodic return for each equivalent goal. For example, in the \textit{ANYmal-Climb} task, we compare the episode returns for the goal of climbing a box forward and backward. For symmetric policies, the variation between the obtained returns for each goal should be low. \tabref{tab:reward-var} shows that for all the tasks, policies trained with augmentation consistently achieve higher average returns while having much lower variation in the returns between symmetric versions of the task. This result shows that learning with symmetry augmentation does lead to more optimal and symmetrical behaviors.

\subsection{Qualitative Behavior Analysis}
\label{sec:qualitative-behavior}

Finally, we describe the different behaviors learned by the policies for all tasks. We refer the reader to the supplementary video for more details.

\subsubsection{CartPole}
Even for this relatively simple task, the behavior of policies trained without augmentation depends on where the pole is initialized. When the pole starts flat on the right side, the policy immediately moves the cart to spin the pole upwards. For the same position on the left, the policy lets it swing towards the other side first, leading to sub-optimal task returns. Policies trained with augmentation exhibit equally optimal behavior from both sides.

\begin{table}[t]
    \centering
    \begin{tabular}{r | rr | rr}
        \toprule
        \multirow{2}{*}{Environment} & \multicolumn{2}{c|}{Vanilla-PPO} & \multicolumn{2}{c}{PPO + aug.} \\
        & Return & Variation & Return & Variation \\
        \midrule
        CartPole & -2.507 & 0.353 & \bf{-1.928} & \bf{0.003}  \\
        \rowcolor[HTML]{EFEFEF} ANYmal-Climb &  15.544 & 1.022 & \bf{17.462} & \bf{0.124} \\
        ANYmal-Push & 16.331 & 2.255 & \bf{18.373} & \bf{0.424} \\
        \rowcolor[HTML]{EFEFEF} Trifinger-Repose & {2153.343} & {75.752}& \bf{2285.125} & \bf{7.884} \\
        \bottomrule
    \end{tabular}
    \caption{We take a set of equivalent goals for each task and report the average episodic returns over 500 runs for each goal. The variation is the maximum difference in the returns between equivalent goals. Since rewards are symmetric, a higher variation implies less symmetric behavior between equivalent goals.
    }
    \vspace{-15.2pt}
    \label{tab:reward-var}
\end{table}

\begin{figure}[t]
    \centering
    \hfill
    \begin{subfigure}[b]{0.47\linewidth}
         \caption{Vanilla PPO}
         \includegraphics[width=0.47\linewidth,trim={0 5.5cm 5.5cm 0}, clip]{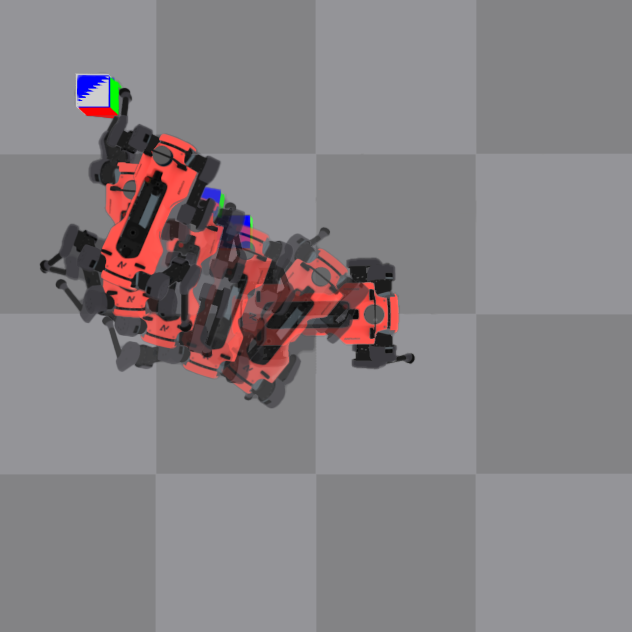}
        \includegraphics[width=0.47\linewidth,trim={5.5cm 5.5cm 0 0}, clip]{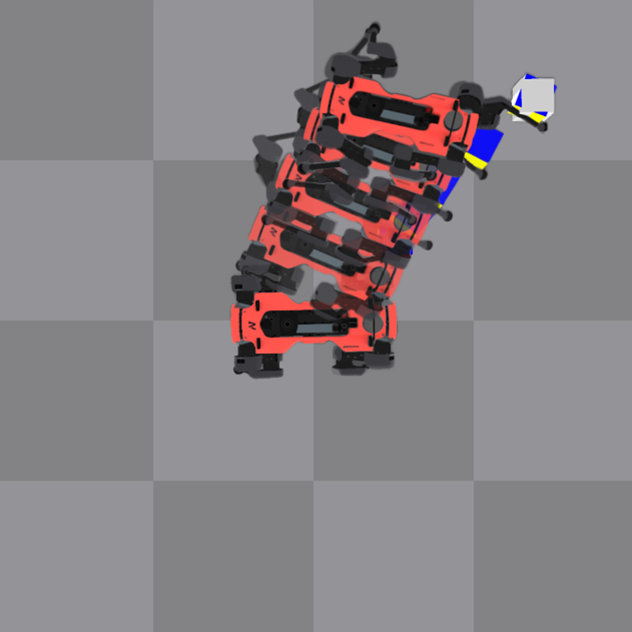}
    \end{subfigure}
    \hfill
    \begin{subfigure}[b]{0.47\linewidth}
        \caption{PPO + aug.}
         \includegraphics[width=0.47\linewidth,trim={0 5.5cm 5.5cm 0}, clip]{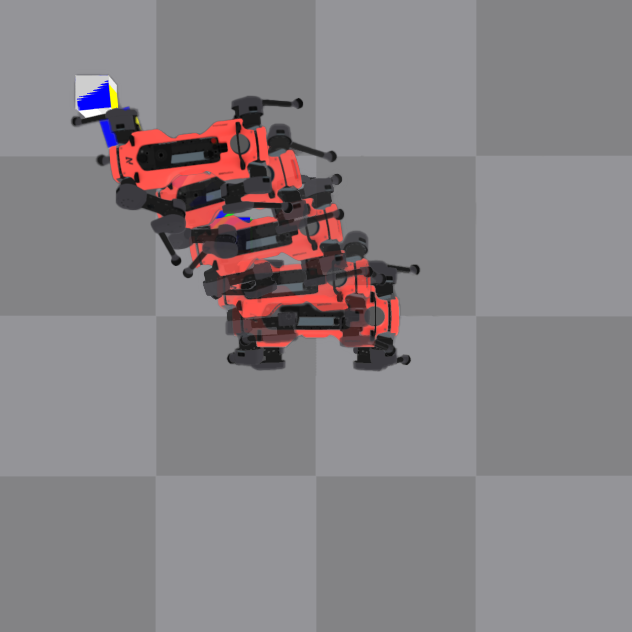}
        \includegraphics[width=0.47\linewidth,trim={1.35cm 1.35cm 0 0}, clip]{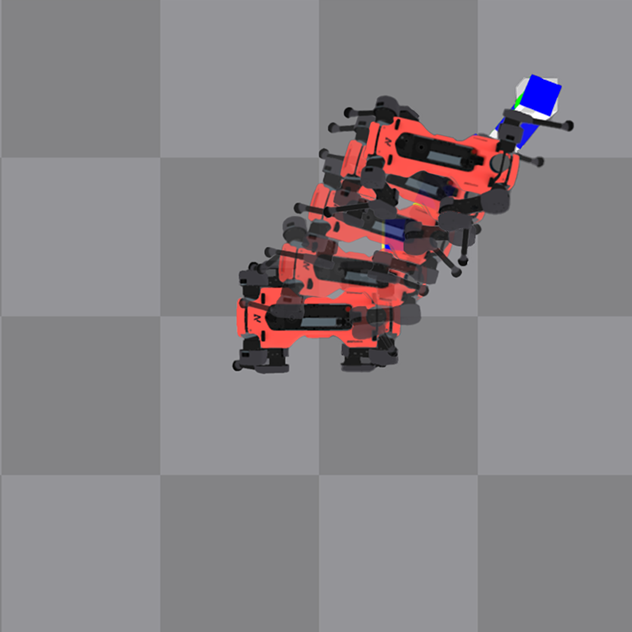}
    \end{subfigure} 
    \\
    \hfill
    \begin{subfigure}[b]{0.47\linewidth}
        \includegraphics[width=0.47\linewidth,trim={0 0 5.5cm 5.5cm}, clip]{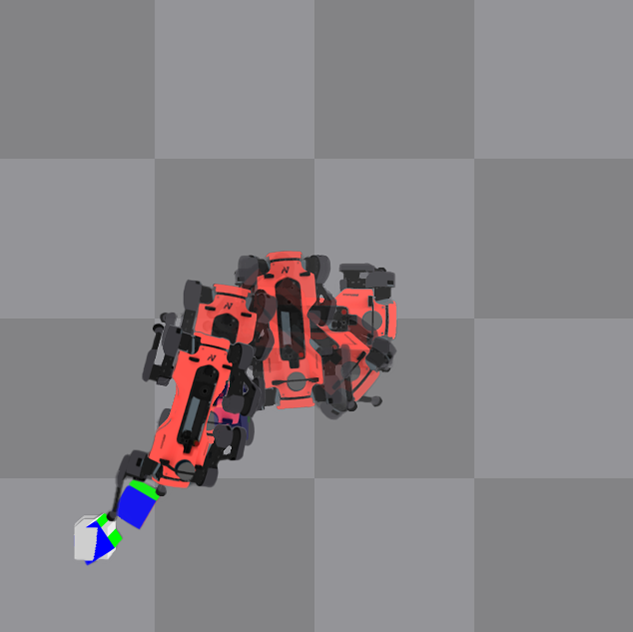}
        \includegraphics[width=0.47\linewidth, trim={5.5cm 0 0 5.5cm}, clip]{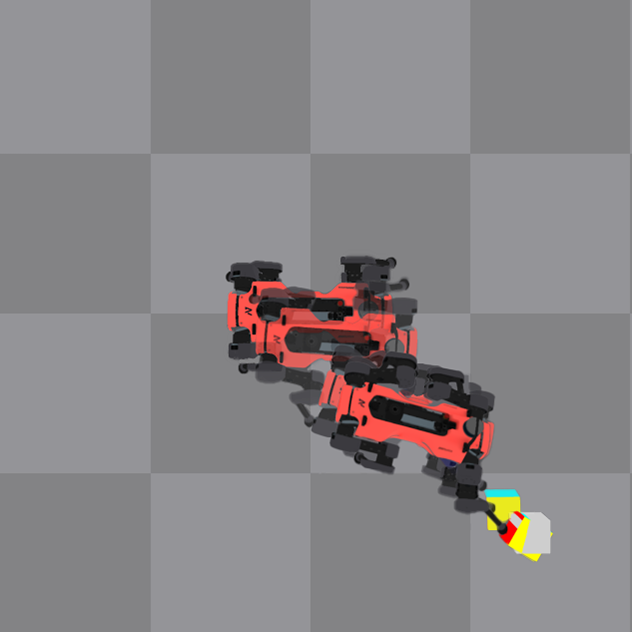}
    \end{subfigure}
    \hfill
    \begin{subfigure}[b]{0.47\linewidth}
         \vspace{2pt}
        \includegraphics[width=0.47\linewidth, trim={0 0 2.75cm 2.75cm}, clip]{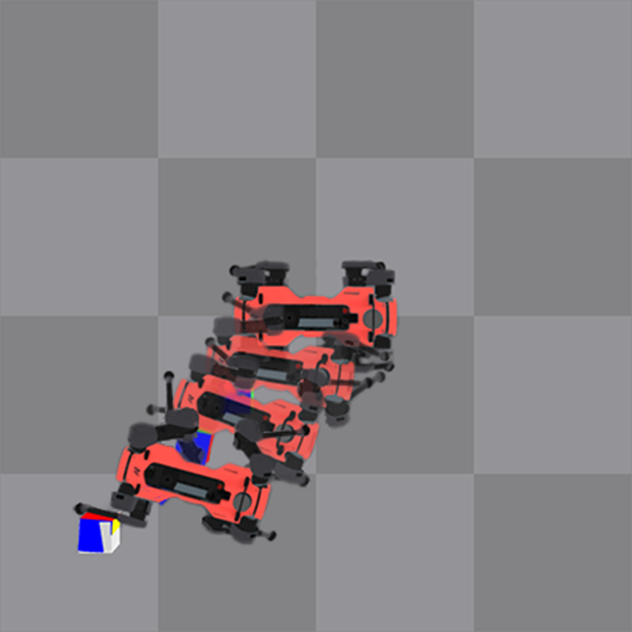}
        \includegraphics[width=0.47\linewidth, trim={1.35cm 0 0 1.35cm}, clip]{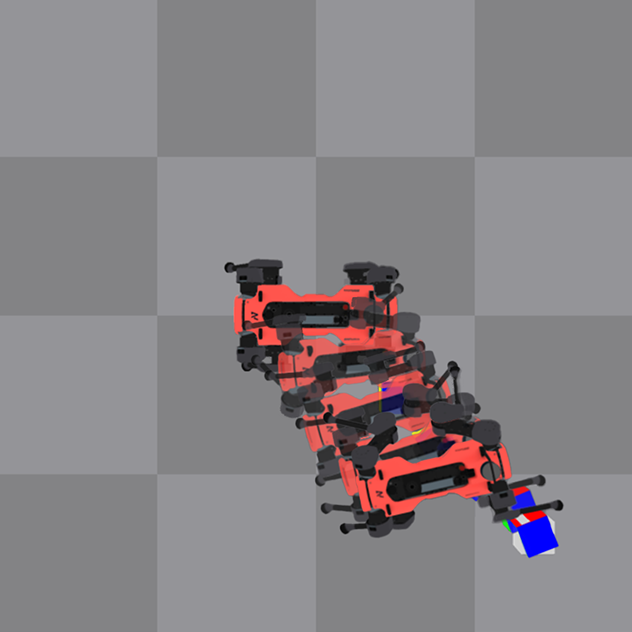}
    \end{subfigure}

    \caption{Observed trajectories for equivalent goals in the ANYmal-Push task. Using data augmentation, the behavior is more symmetrical and the robot uses all of its legs for manipulation.}
    \label{fig:learned-motions-sim}
    \vspace{-1.5pt}
\end{figure}

\subsubsection{ANYmal-Climb}

\begin{figure}[t]
    \centering
    \begin{minipage}[b]{0.95\linewidth}

        \includegraphics[width=\linewidth]{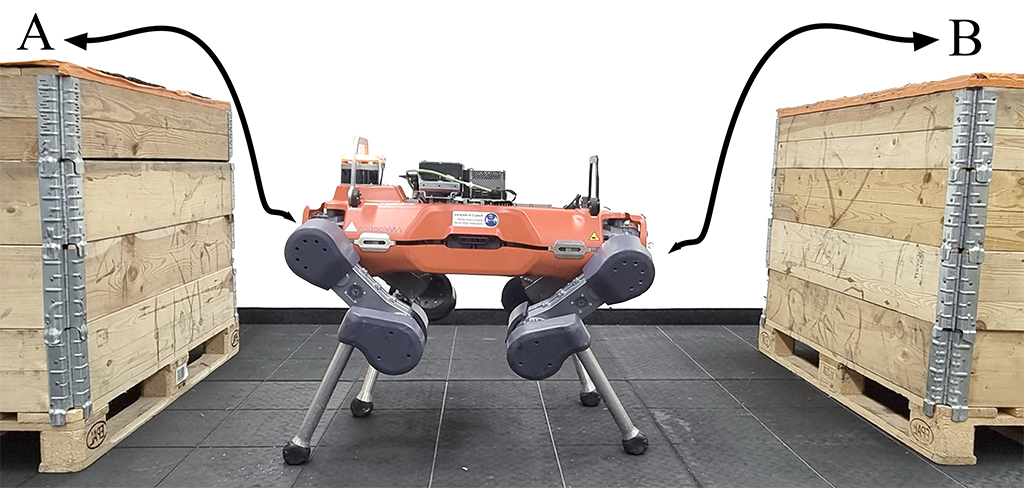}
        \begin{tikzonimage}[width=0.325\linewidth,trim={5cm 0 10cm 0},clip]{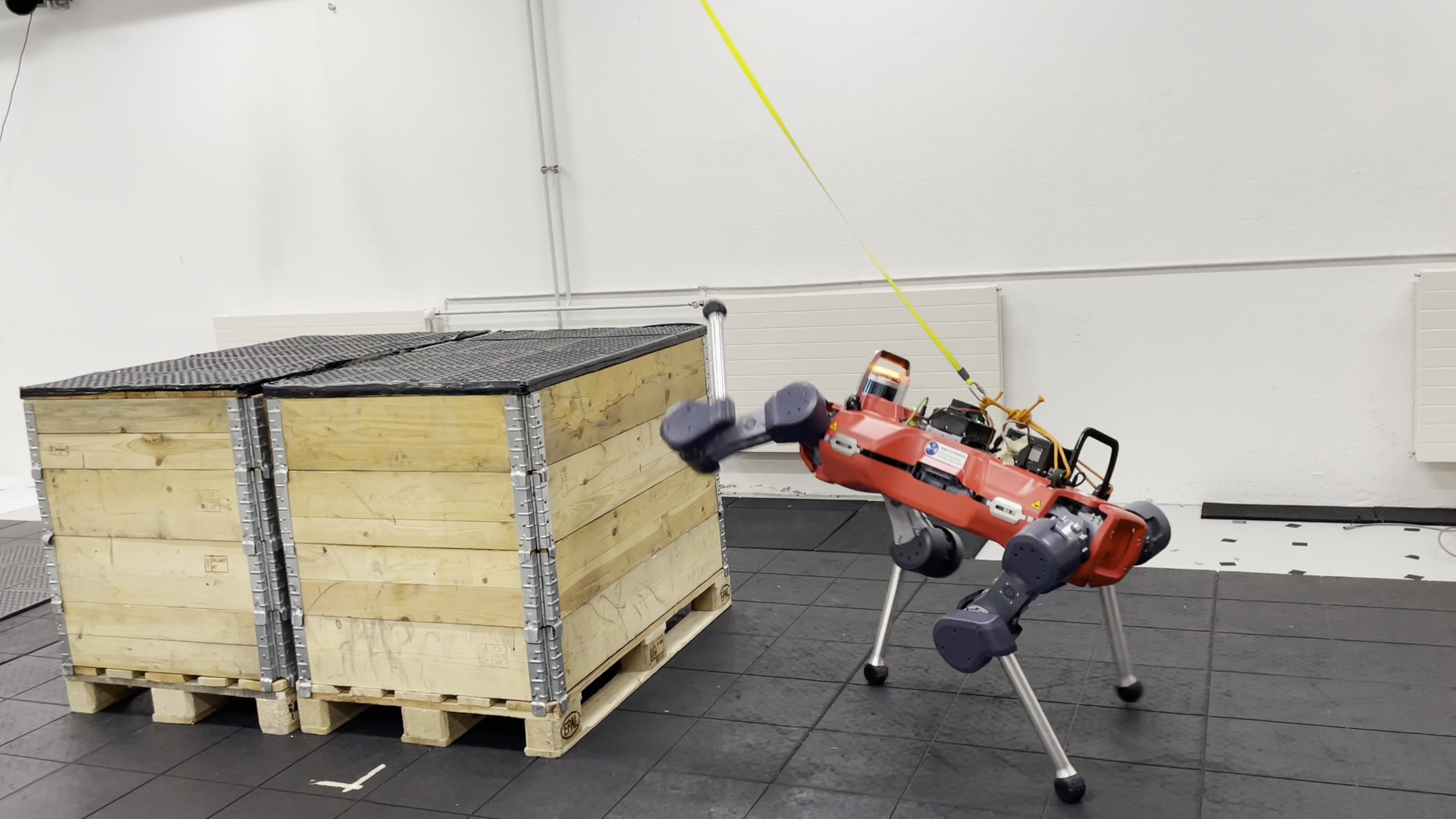}[image label]
            \node{A.1};
        \end{tikzonimage}
        \begin{tikzonimage}[width=0.325\linewidth,trim={5cm 0 10cm 0},clip]{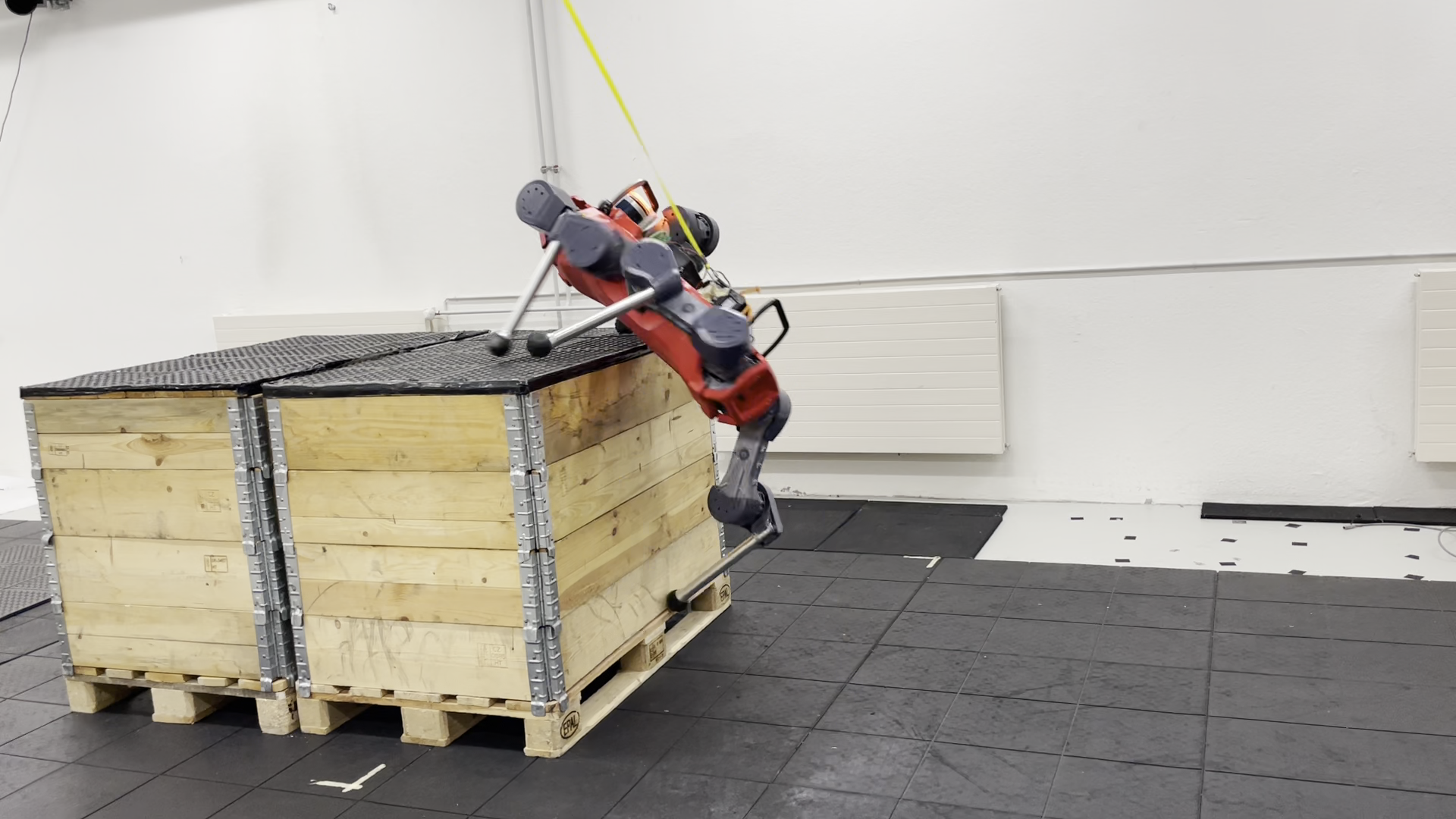}[image label]
            \node{A.2};
        \end{tikzonimage}
        \begin{tikzonimage}[width=0.325\linewidth,trim={5cm 0 10cm 0},clip]{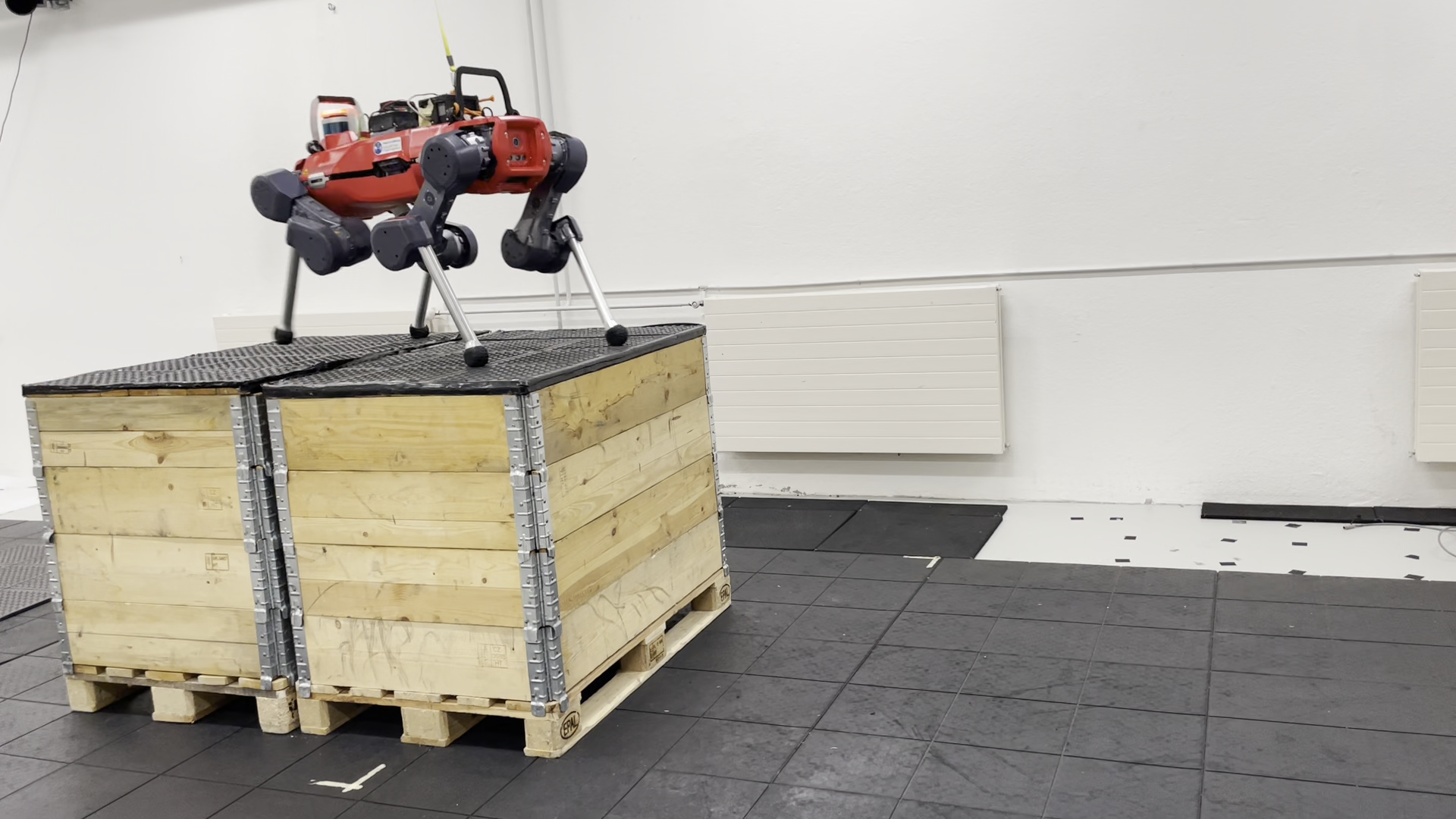}[image label]
            \node{A.3};
        \end{tikzonimage}

    \end{minipage}
    \caption{Hardware deployment for the ANYmal-Climb task. The panel below shows the execution of the policy trained with symmetry augmentation to reach $A$. Please check the supplementary video for comparisons with behaviors obtained using vanilla-PPO.}
    \label{fig:hardware-test}
    \vspace{-2.5pt}
\end{figure}

Policies trained without augmentation usually learn to climb only in one direction, always using the same leg first. When the robot is initialized in another direction, the policy prefers to turn on the spot before climbing. Since we set the initial and target orientations as the same, the policy turns again on the box to reorient itself. This leads to sub-optimal policies that turn twice instead of directly climbing backward.
Training with augmentation mitigates this issue and the policies can climb forward and backward while using any of the legs to initiate the climbing.

\subsubsection{ANYmal-Push}

In this loco-manipulation task, the asymmetry in the learned policy with vanilla-PPO is more prominent since the robot uses only some of its legs for walking while the others for manipulating the object. Regardless of the uniform sampling of the object and its target around the robot, policies typically push the object with only two of its limbs and turn around to use only those two limbs for manipulation (\figref{fig:learned-motions-sim}). With symmetry augmentation, the robot uses all the limbs depending on whichever is closest to the object. It does this without any hand-crafted rewards to encourage a certain end-effector to move towards the object.

\subsubsection{Trifinger-Repose}

The policies trained without augmentation learn different finger gaits for rotationally equivalent goals. For example, the robot may flip the cube on the table before picking it up, while sometimes directly picking it up. In contrast, policies trained with augmentation produce the same pattern for rotationally similar goals and also complete the task faster.

\subsection{Hardware Deployment}

We conduct hardware deployment on ANYmal-D for the ANYmal-Climb task (\figref{fig:hardware-test}). We find that policies from vanilla-PPO result in fast re-orienting behaviors that often cause perception failures and missteps. In contrast, policies trained with augmentation avoid these unnecessary rotations and display more predictable and robust behaviors. It is worth highlighting that even though the real robot is not perfectly symmetrical (uneven payload and wear-and-tear of the actuators), the policies trained with augmentation are resilient to these asymmetries and achieve successful box climbing maneuvers. One possible explanation for this success lies in the approach's emphasis on encouraging symmetry while allowing the policy to adapt naturally to the robot's asymmetries during training.

\section{Discussion}

We investigated two approaches for inducing symmetry invariance in on-policy DRL methods for goal-conditioned tasks. We presented an alternate update rule for symmetry-based data augmentation that helps stabilize the learning in practice. We compared the two approaches on various robotic tasks and showed how data augmentation leads to faster convergence with virtually symmetric and more optimal policies. Through hardware deployment for the quadrupedal agile locomotion task, we demonstrated that the policy learned with data augmentation transfers well even when the hardware is not perfectly symmetrical.

While this work mathematically motivates and empirically justifies the importance of initializing with small weights for data augmentation, a more rigorous treatment is for future work. Further investigation is also needed to understand how to perform augmentation when the symmetry in the MDP and the transformations are not explicitly available. For instance, for the latent vector obtained from an autoencoder. 

\bibliographystyle{IEEEtranBST/IEEEtran} %
\bibliography{IEEEtranBST/IEEEabrv,root}  %

\end{document}